\documentclass[10pt,twocolumn,letterpaper]{article}
\usepackage{cvpr}
\usepackage{times}
\usepackage{epsfig}
\usepackage{graphicx}
\usepackage{amsmath}
\usepackage{amssymb}
\usepackage{amsthm,mathrsfs,amsfonts,dsfont}
\usepackage{float}
\usepackage{multirow}
\usepackage{longtable}
\usepackage{supertabular}
\usepackage{rotating}
\usepackage{subfigure}
\usepackage[breaklinks=true, colorlinks, bookmarks=false]{hyperref}

\cvprfinalcopy 

\def\cvprPaperID{****} 

\begin{document}

\title{SEP-Nets: Small and Effective Pattern Networks}

\author{Zhe Li$^{1,2}$~~~~Xiaoyu Wang$^2$~~~~Xutao Lv$^2$~~~~Tianbao Yang$^1$\\
$^1$The University of Iowa~~~~~~~~~~~~~~~~$^2$Snap Research\\
{\tt\small \{zhe-li-1,tianbao-yang\}@uiowa.edu~~~fanghuaxue@gmail.com~~~xutao.lv@snap.com}
}

\maketitle

\begin{abstract}
   While going deeper has been witnessed to improve the performance of convolutional neural networks (CNN), going smaller for CNN has received increasing attention recently due to its attractiveness for mobile/embedded applications. It remains an active and important topic how to design a small network while retaining the performance of large and deep CNNs (e.g., Inception Nets, ResNets). Albeit there are already intensive studies on compressing the size of CNNs, the considerable drop of performance is still a key concern in many designs. This paper addresses this concern with several new contributions. First, we propose a simple yet powerful method for compressing the size of deep CNNs based on parameter binarization. The striking difference from most previous work on parameter binarization/quantization lies at different treatments of $1\times 1$ convolutions and $k\times k$ convolutions ($k>1$), where we only binarize $k\times k$ convolutions into binary patterns.  The resulting networks are referred to as pattern networks. By doing this, we show that previous deep CNNs such as GoogLeNet and Inception-type Nets can be compressed dramatically with marginal drop in performance. Second, in light of the different functionalities of $1\times 1$ (data projection/transformation) and $k\times k$ convolutions (pattern extraction), we propose a new block structure codenamed the pattern residual block that adds transformed feature maps generated by $1\times 1$ convolutions to the pattern feature maps generated by $k\times k$ convolutions, based on which we design a small network with $\sim 1$ million parameters.  Combining with our parameter binarization, we achieve better performance on ImageNet than using similar sized networks including recently released Google MobileNets.
\end{abstract}

\section{Introduction}
Deep convolutional neural networks  have already achieved tremendous success on a variety of computer vision tasks such as image classification \cite{Alexnet12,vggnet15,googlenet15,ResNet_cvpr16}, object detection \cite{RCNN14,sermanet2013overfeat,fastrcnn15,ren2015faster}, segmentation \cite{long2015fully,he2017mask}, video analysis \cite{xu2015discriminative,zha2015exploiting,gan2015devnet}, human pose estimation \cite{toshev2014deeppose}  among many others. The performance on these different tasks are dramatically boosted  by  sophisticated neural network structures such as AlexNet \cite{Alexnet12}, NIN (Network In Network) \cite{lin2013network}, VGG-Net \cite{vggnet15}, Inception Network \cite{googlenet15}, and ResNet \cite{ResNet_cvpr16}. It is clear that  the networks are going deeper and deeper from AlexNet to ResNets. 

On the other hand, due to the need in mobile/embedded applications, there is a new trend of going smaller while retaining the performance of large and deep CNNs. While Inception Nets and ResNets have tried to reduce the model size by reducing the size of convolution kernels,  using $1\times 1$ convolutions and trimming the fully connected layers, they are still too large to meet the demanding requirement for mobile and embedded devices (e.g., FPGA). For example, ResNet-101 has 200MB and GoogLeNet has 50MB. However, FPGAs often have less than 10MB  of on-chip
memory and no off-chip memory or storage~\cite{iandola2016squeezenet}. 
To further reduce the model size, various compressing techniques have been introduced to deep CNNs, including parameter quantization, binarization, sharing, pruning, hashing, Huffman coding, etc~\cite{cai2017deep,courbariaux2016binarized,han2015adeep,han2015learning,chen2015compressing,hubara2016binarized,courbariaux2015binaryconnect,rastegari2016xnor,ott2016recurrent,zhou2016dorefa,lin2016overcoming}. 
There also emerge few studies recently attempting to design small and compact networks, including the SqueezeNets~\cite{iandola2016squeezenet} and the MobileNets~\cite{howard2017mobilenets}. Nevertheless, the performance drop of smaller networks is still a critical concern for many designs. For example, the authors of~\cite{iandola2016squeezenet} have designed an extremely small network with less than 0.5MB and achieved 57.5\% top-1 accuracy on ImageNet, which is considerably  less than state-of-the-art results of deep CNNs (e.g., 68.65\% of GoogLeNet according to our implementation).

In this paper, we address this concern by proposing several new techniques in the two aforementioned directions for reducing the model size. {\bf First}, we consider parameter binarization - a simple and effective method for reducing the model size. While many previous works try to quantize or binarize all weights in deep CNNs, we propose a novel treatment of $1\times 1$ kernels and $k\times k$ kernels (e.g., $k=3, 5$). In particular, we only binarize $k\times k$ convolutional kernels (with $k>1$). This design is motivated by the difference between $1\times 1$ convolutions and  $k\times k$ convolutions and  the community’s prior knowledge about them. Unlike $k\times k$ convolutions that explicitly extract features in a spatial manner, $1\times 1$ convolutions serve as data projection and transformation. In this sense, $1\times 1$ convolutions need to preserve the information as much as possible and $k\times k$ convolutions is only required to extract abstract patterns from images. In addition,  many works in computer vision have used binary convolutions to extracted features from images~\cite{whitehill2006haar,st2016fast}, while sparse projection has been reported with performance drop compared with dense projection~\cite{xu2016efficient}. The different treatments of $1\times 1$ and $k\times k$ kernels also has several benefits in terms of computation: (i) $1\times 1$ convolutions using floating points is cheaper and simpler than $k\times k$ convolutions; (ii) this splitting is very suitable for FPGAs where logic blocks can efficiently handle the binarized convolutions and DSP units can handle the $1\times 1$ convolutions. {\bf Second}, we propose a simple new design of small networks by stacking up several layers of a novel module, which is built on a new block codenamed pattern residual block. The idea of the pattern residual block is to add transformed feature maps generated by $1\times 1$ convolutions to the pattern feature maps generated by $k\times k$ convolutions, which resembles but generalizes the skip connection in ResNets. 
The new pattern residual block is well suited to the design of small networks for increasing the model capacity and more importantly to the binarized pattern networks for offsetting the effect of pattern binarization. Using 5.2MB, our designed small network (termed as \textbf{SEP-Net}) achieves $65.8\%$ top-1 accuracy, beating that of the SqueezeNet (4.8MB, 60.4\%) and the MobileNet (5.2MB, 63.7\%) with simlar sizes. Leveraging our pattern binarization, we  reduce our model size to 4.2MB while maintaining 63.7\% top-1 accuracy. By further quantizing $1\times 1$ filters using $8$ bits,  we achieve 63.5\% top-1 accuracy with a model size  1.3MB. 

\section{Related Work}\label{related-work}
In this section, we briefly review the related work on designing modern network structures and techniques for reducing neural network model size. 
Since introduced in~\cite{lin2013network}, the $1\times 1$ convolutions have been extensively used in modern networks such as Inception Nets and ResNets, which can reduce the number of parameters comparing with large convolutional kernels. In these network designs, the $1\times 1$ convolutions mainly serve as data projection for reducing the channels of feature maps. Inception Nets also concatenate the $1\times 1$ convolved feature maps and the $k\times k$ convolved feature maps in the inception modules. In this paper, we will innovatively leverage the power of $1\times 1$ convolutions to improve the performance of binarized  networks.  In addition, we will utilize the data transformation capability of $1\times 1$ convolutions to design a new residual block. 

Recently, there emerge intensive studies on compressing the size of CNNs. Various techniques have been introduced to CNNs to either reduce the number of parameters or reduce the size of parameter representations. These include weight pruning \cite{han2015deep,han2015learning}, weight binarization~\cite{hubara2016binarized,courbariaux2015binaryconnect,rastegari2016xnor}, weight ternarization~\cite{ott2016recurrent,yin2016training,mellempudi2017ternary,li2016ternary,lin2015neural}, weight quantization \cite{courbariaux2016binarized,cai2017deep} and designing small and compact networks \cite{lin2013network,iandola2016squeezenet}. 
There are several differences between the proposed weight binarization and previous work on weight binarization~\cite{hubara2016binarized}. First, unlike previous work that binarize all weights,  we only binarize the weights of $k\times k$ filters ($k>1$). Second, our focus is not to reduce the computational costs of training by binarization but instead to reduce the costs for deploying the model, which is clearly different from some previous works focusing on training, e.g., BinaryConnect~\cite{courbariaux2015binaryconnect}, Binarized Neural Networks~\cite{hubara2016binarized}, XNOR-Nets~\cite{rastegari2016xnor}. Our approach is to directly binarize fully trained deep CNNs and fine-tune the $1\times 1$ filters. The benefit of this two-step approach is that it will not suffer from difficulties of training binarized networks and therefore enjoy less performance drop. 

The design of small and compact networks is the focus of several recent works. Xu et al.~\cite{juefei2016local} exploited the local binary convolutional operators in deep CNNs.  They utilize traditional local binary operators in place of $k\times k$ ($k>1$) convolutions.  The difference from our work  is that the binary convolutional filters in their work are randomly generated and fixed during training of the networks. In addition, the performance of their networks on large scale ImageNet data is not shown. The SqueezeNet explored several strategies to reduce the number of parameters including (i) replacing $k\times k$ convolutions ($k>1$) by $1\times 1$ convolutions; (ii) decreasing the number of input channels to $3\times 3$ filters; and (iii) postponing the down sampling to late layers in the network~\cite{iandola2016squeezenet}. They designed Fire module that consists of $1\times 1$ convolutional layers to squeeze the size of feature maps and an expand layer  that has a mix of 1x1 and 3x3 convolution filters. The MobileNets approximate the standard $k\times k$ ($k>1$) convolutions by depth-wise convolutions and $1\times 1$ convolutions, and also introduce  two  hyper-parameters to balance between latency and accuracy~\cite{howard2017mobilenets}. The small pattern networks explored in this paper share some similarities to these previous small networks in that  much computation burden will shift to the $1\times 1$ convolutions but also bear some subtle differences. Most importantly, the present work has achieved the best performance on ImageNet data with the same network size among the MobileNets and SqueezeNets.


\section{Our Approach}\label{approach}
In this section, we present the crucial ingredients of the designed \textbf{SEP-Net}: Pattern Binarization and Pattern Residual Block. It is notable that these two techniques are of their own interest and could be employed in design of other small neural network structures.  Following those techniques, we present the detailed architecture of the designed \textbf{SEP-Net}.

\subsection{Pattern Binarization}
\begin{figure}[t]
\centering
\includegraphics[scale=0.35]{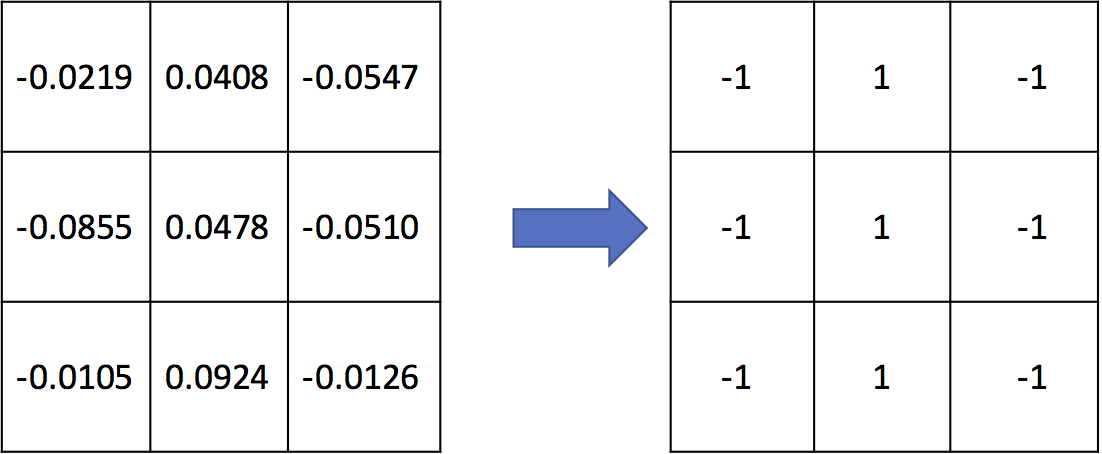}
\caption{A trained $3\times 3$ filter from  GoogLeNet (Left), and its binarized version (Right)}
\label{fig:binaryfilter-sep-net-detail}
  \end{figure}

Since fully connected layers have been removed in modern deep CNNs (including Inception Nets, ResNets), here we only consider parameterized convolutional layers. 
We adopt the following simple procedure to obtain a compressed network from any successful network structures including our designed SEP-Net as described later: 
\begin{itemize}
\item Step 1: train a full neural network such as GoogLeNet, ResNet and SEP-Net from scratch.
\item Step 2: binarize $k\times k$ ($k>1$) convolutional filters in the well-trained neural network model. 
\item Step 3: fine-tune the scaling factors of all binarized $k\times k$ filters and the floating point representations of all $1\times 1$ filters by back-propagation on the same dataset. 
\end{itemize}
The different treatments of $1\times 1$ filters and $k\times k$ filters is motivated by their complementary roles in CNNs.  $k\times k$ filters serve as spatial pattern extraction from an input image/feature map, while $1\times 1$ filters mainly serve as data projection and transformation. To justify our choice,  we also quantitatively analyze the effect of binarizing $1\times 1$ filters and $k\times k$ filters from the viewpoint of quantization error. In particular, if we let $W$ denote an $c\times k\times k$ convolutional filter. The binarization seeks to approximate it by $\alpha B$, where $B$ is a binary filter with entries from $\{1,-1\}$ and $\alpha$ is a scaling factor. From the viewpoint of minimizing the quantization error, $\alpha, B$ can be sought by solving the following problem:
\begin{align}
\min_{\alpha\in\mathbb R, B\in\{1,-1\}^{c\times k\times k}} E(W, B, \alpha)\triangleq  \|W - \alpha B\|_F^2
\end{align}
The optimal solutions have been studied in~\cite{iandola2016squeezenet}. Actually, the optimal $B^*$ can be found by thresholding, i.e., $B^*_{i,j,l}=1$ if $W_{i,j,l}\geq 0$ and $B^*_{i, j, l}=-1$ if $W_{i, j, l}<0$. This binarization procedure is illustrated in Figure~\ref{fig:binaryfilter-sep-net-detail}.  The optimal $\alpha_*$ can be computed by $\alpha_* = \frac{\sum_{i,j, l}|W_{i,j,l}|}{c\times k\times k}$. To quantitatively understand the effect of binarizing $1\times 1$ filters and $k\times k$ filters, we first train a fully GoogLeNet \cite{googlenet15},  which is composed of $1\times 1$, $3\times 3$ and $5\times 5$ convolutions filters. 
Then we compute the quantization error for all filters, and obtain averaged quantization error for $1\times 1$, $3\times 3$ and $5\times 5$ filters, respectively. The result is reported in Table~\ref{tab:qe}, from which we can clearly see that that quantization error for $1\times 1$ filters is by a order of magnitude larger than that of $3\times 3$ and $5\times 5$ filters. This justifies our choice of binarizing $k\times k$ filters ($k>1$) while retaining $1\times 1$ filters. We present more results on prediction performance by binarizing $1\times 1$, $3\times 3$ and $5\times 5$ filters. 

\begin{table}
\caption{Averaged Quantization Error of Different Sized Filters from GoogLeNet.}\label{tab:qe}
\centering
\begin{tabular}{l|l|l}
  \hline
$1\times 1$& $3\times 3$&$5\times 5$\\\hline
$0.0462$&$0.0029$& $0.0056$\\ \hline
 \end{tabular}
\end{table}

\subsection{Pattern Residual Block and the Architecture of \textbf{SEP-Net}}
\label{sec:PRB}
Pattern binarization is an effective method for compressing the size of a large and deep CNN. We can apply this technique to reducing the size of previous deep CNNs (e.g., GoogLeNets, Inception Nets and ResNets). However, due to that the original sizes of these deep CNNs are very large, the resulting pattern networks from these deep CNNs may not be small enough for deployment in mobile and embedded applications. To address this issue, we propose a new design of a small and effective network  (\textbf{SEP-Net}). 

\begin{figure*}[t]
 \centering
   \subfigure[Pattern Residual Block ]{\includegraphics[scale=0.25]{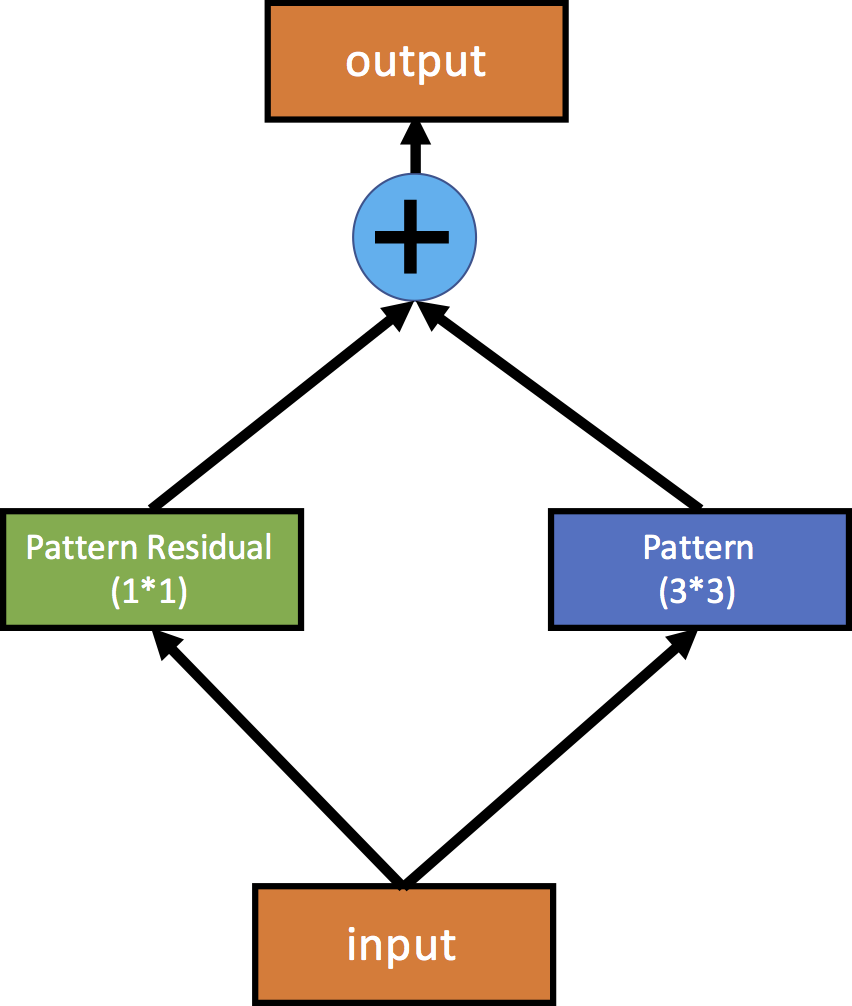}}
   \hspace{1.4in}
   \subfigure[Group-wise convolution]{\includegraphics[scale=0.3]{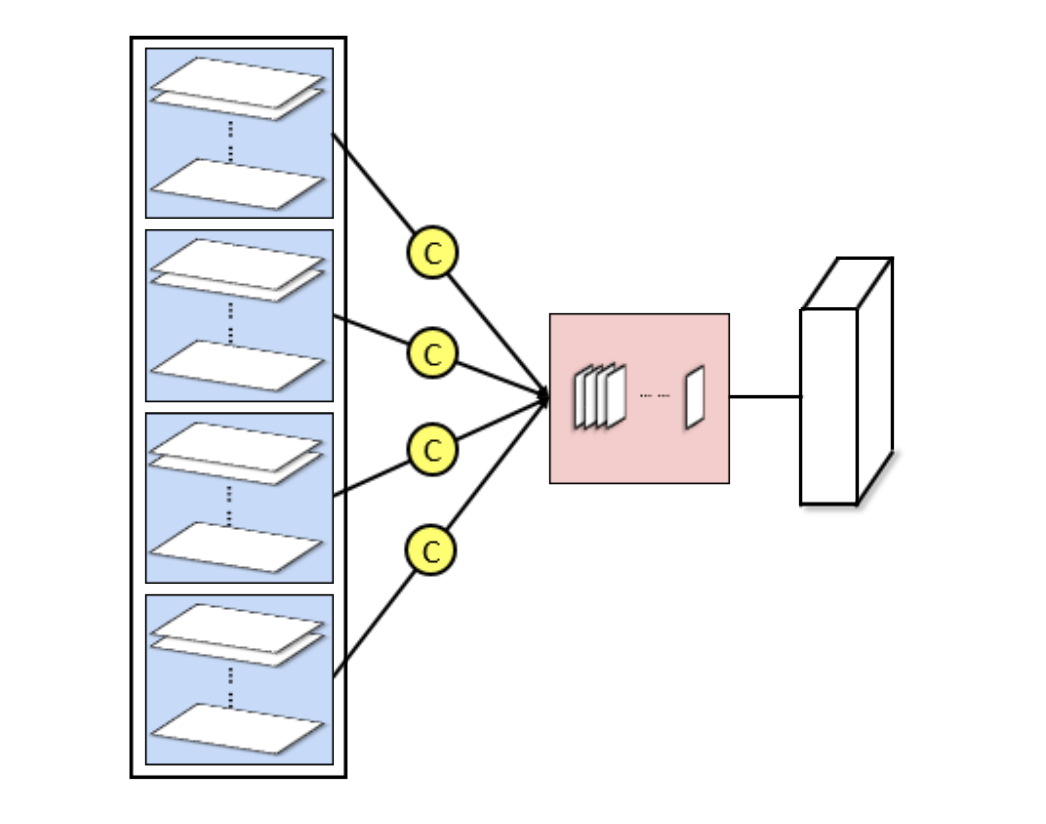}}
\caption{Left: Pattern Residual Block; Right: Illustration of Group-wise convolution}
 \label{fig:group-convolution}
\end{figure*}

\paragraph{Pattern Residual Block.}  We first describe the building block of our design -  a novel block codenamed pattern residual block (PRB). 
As shown in Figure~\ref{fig:group-convolution}(a), the PRB consists of  $1\times1$ convolutions and  $k\times k$ convolutions (in particular $k=3$), which are executed in parallel and their feature maps are added together. In particular, if we let $x$ denote an input, the output of this building block can be expressed as $O(x) = C_{k\times k}(x) + C_{1\times 1}(x)$, where $C_{k\times k}$ denotes a $k\times k$ convolution ($k>1$) and $C_{1\times 1}$ represents $1\times 1$ convolution. If we consider $x$ a vector, $C_{1\times 1}(x)$ is equivalent to $Ax$, where $A$ is a linear mapping. Since this is a generalization of identity mapping, we find it particularly useful in building pattern networks. Especially when the pattern block ($3*3$ convolutions) is binarized, the additive $1\times 1$ convolutions is able to offset the change incurred by the binarization, which acts the residual between fully $3*3$ filtered maps and binarized $3*3$ filtered maps.   

 \begin{figure*}[t]
 \centering
 
 \hspace*{-0.2in}\subfigure[Inception Module ]{\includegraphics[width=0.23\textwidth]{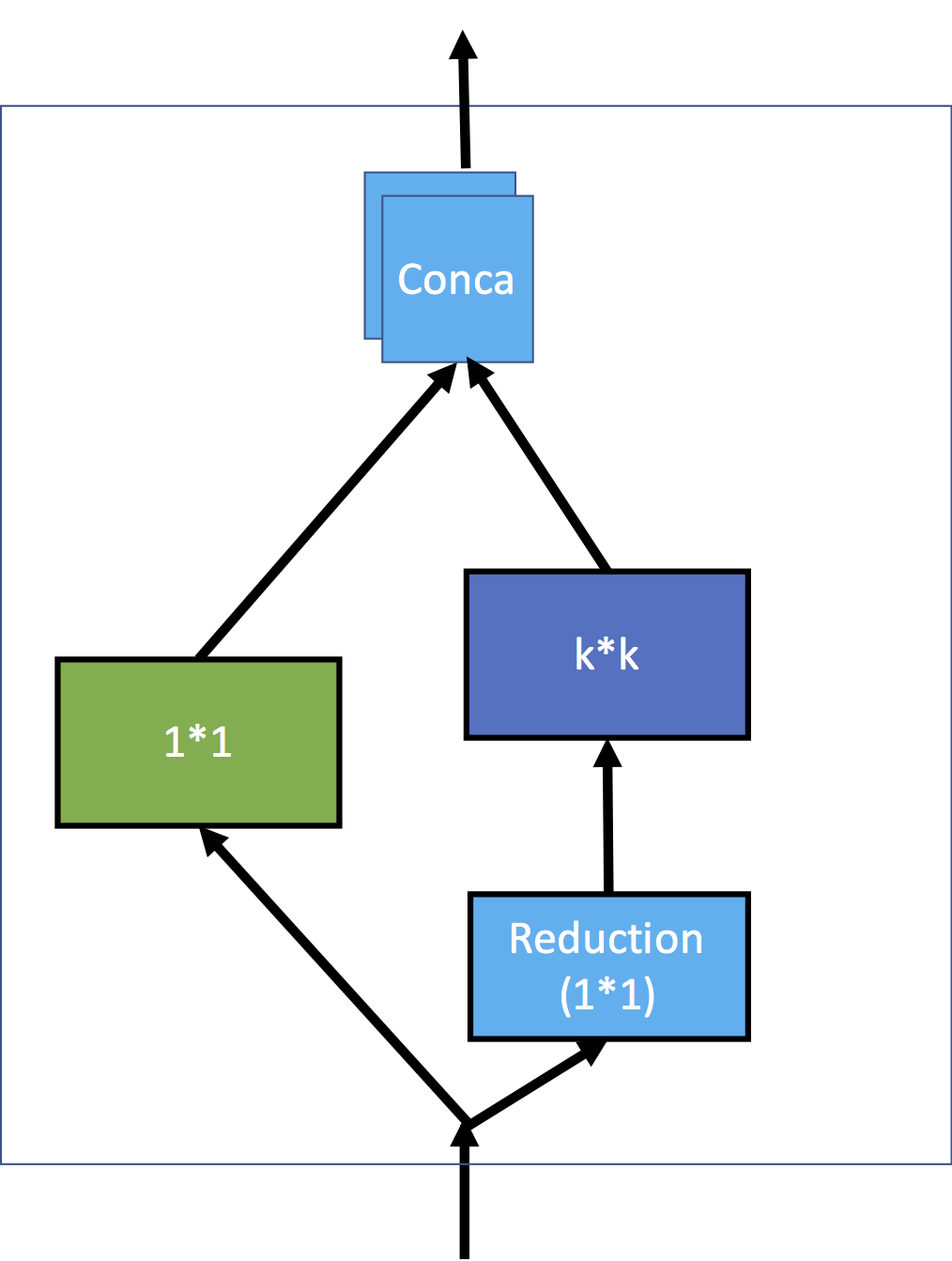}}\hspace*{0.55in}
 \subfigure[Inception-ResNet Module ]{\includegraphics[width=0.26\textwidth]{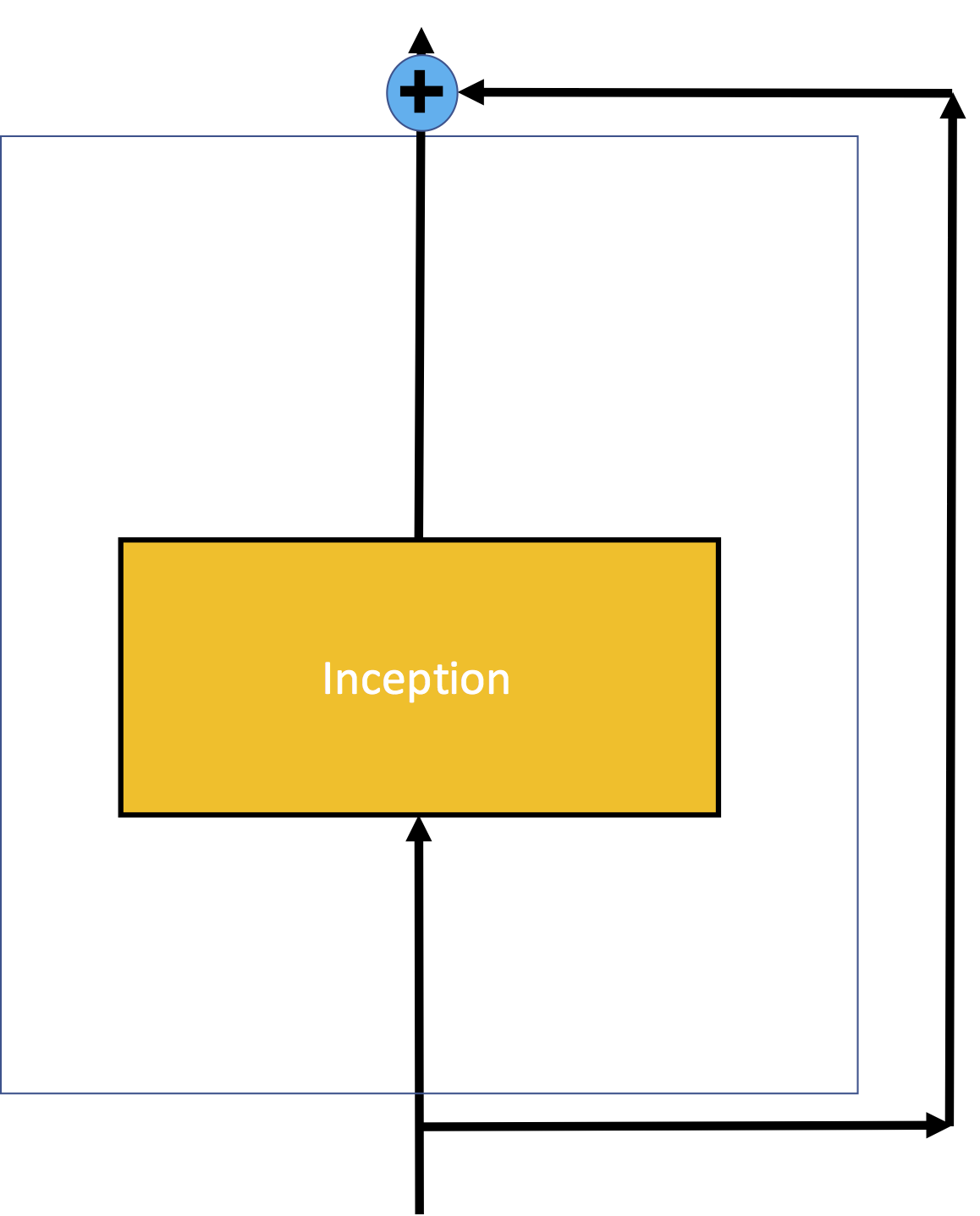}}\hspace*{0.55in}
 \subfigure[the SqueezeNet  Module ]{\includegraphics[width=0.26\textwidth]{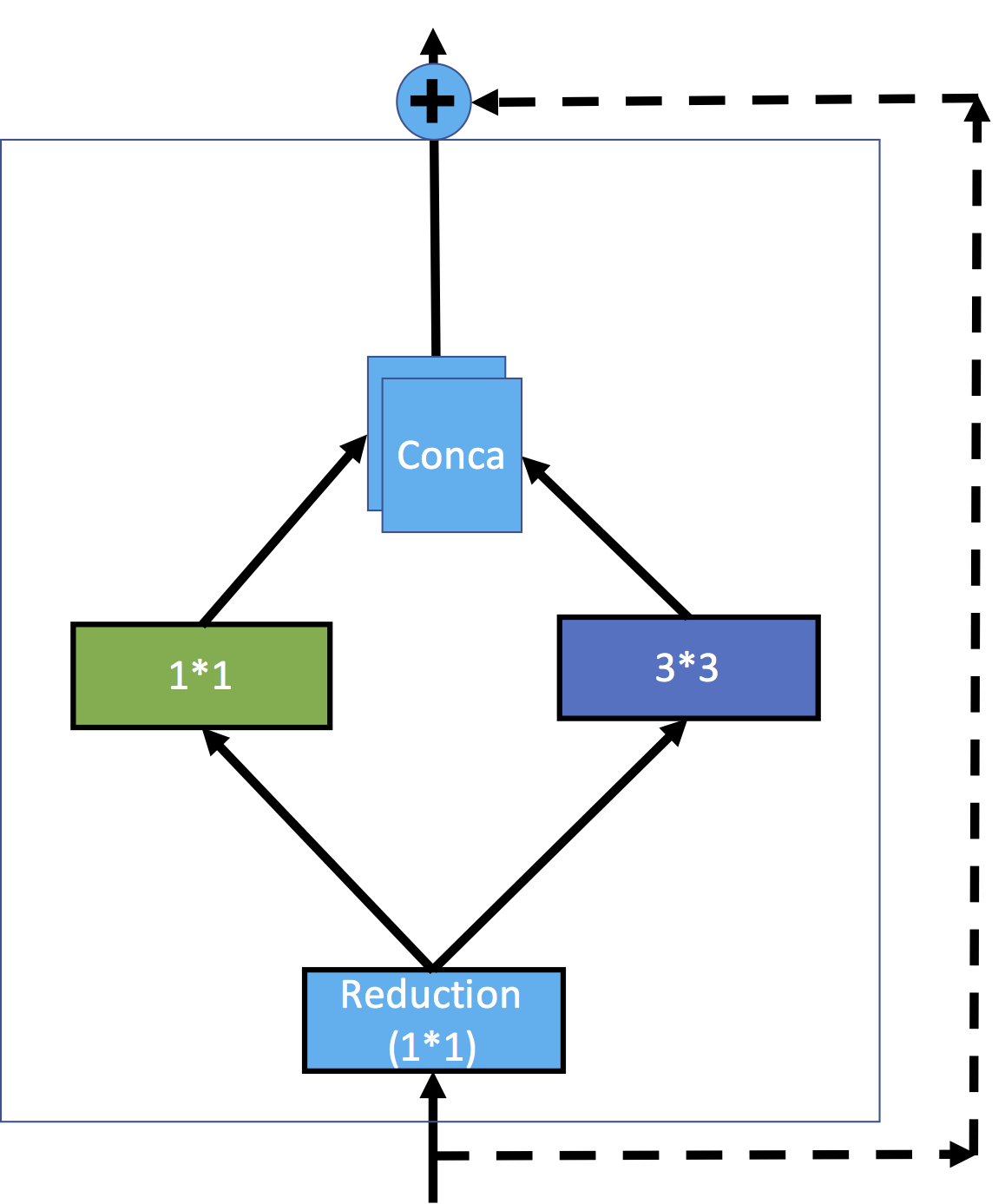}}

    \hspace*{-0.2in}\subfigure[ResNet Module ]{\includegraphics[width=0.25\textwidth]{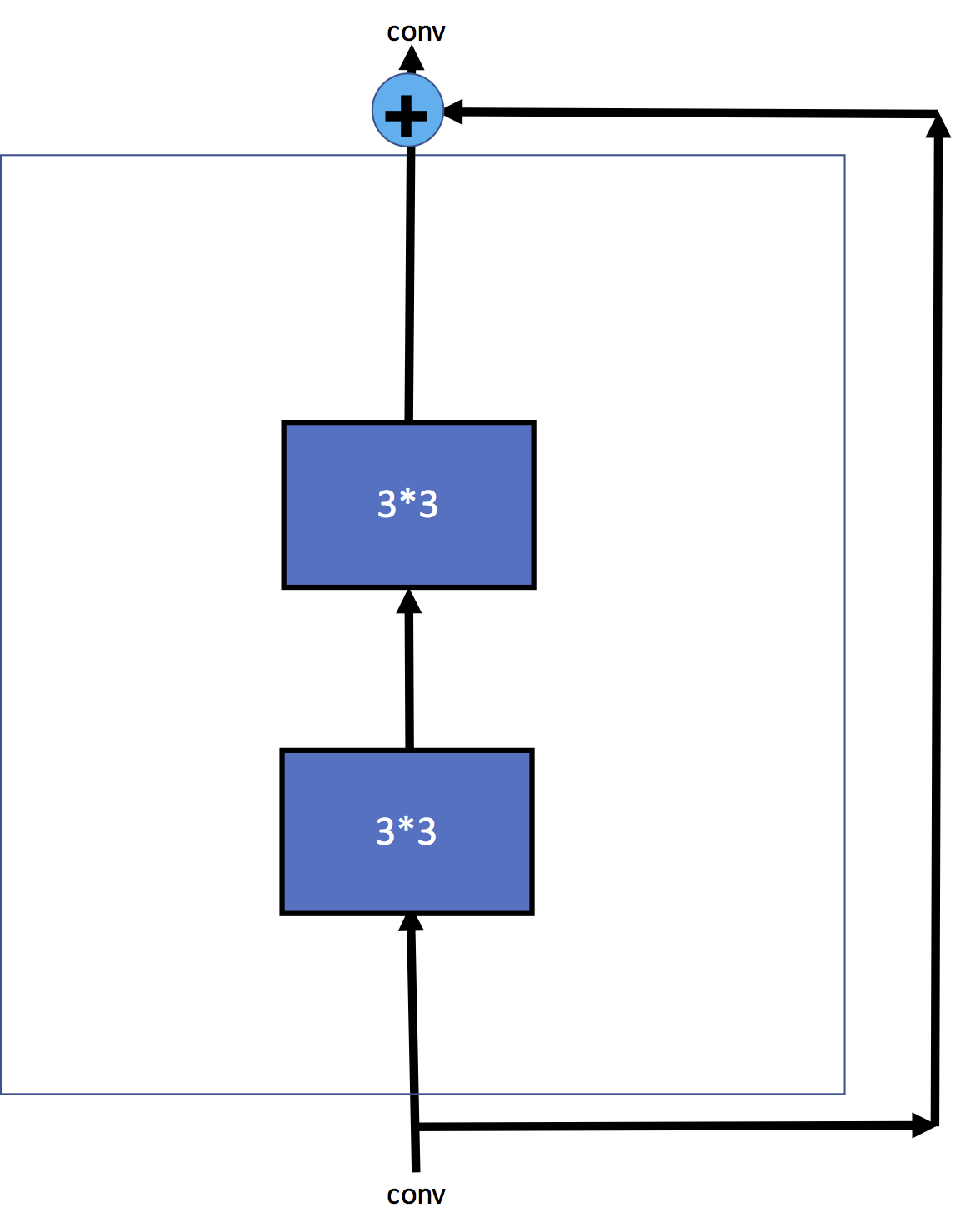}}\hspace*{0.55in}
    \subfigure[ResNet ``bottleneck" Module ]{\includegraphics[width=0.25\textwidth]{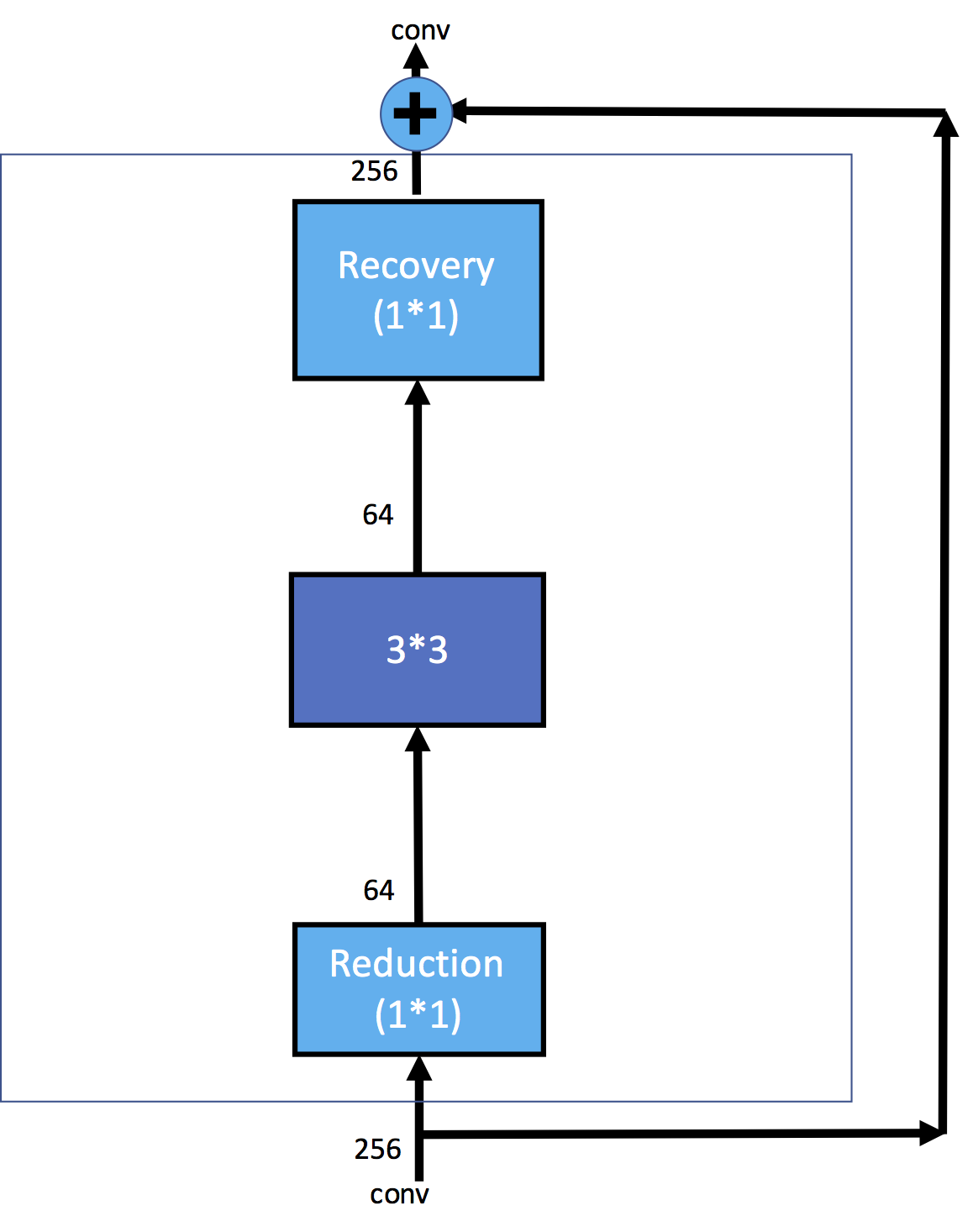}}\hspace*{0.55in}
  \subfigure[SEP-Net Module ]{\includegraphics[width=0.25\textwidth]{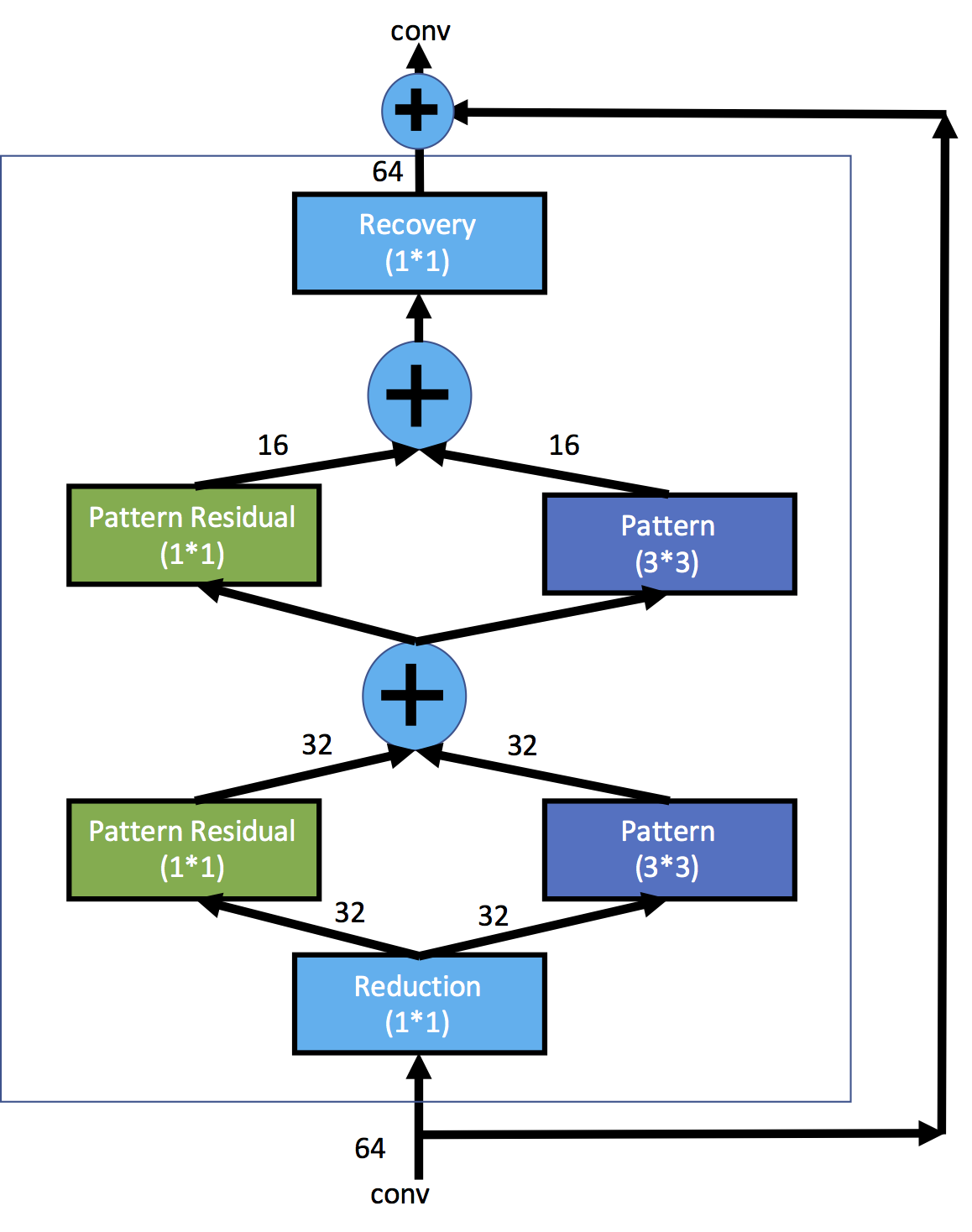}}

    
 \caption{Different Network Modules (dashed line represents either the identity mapping or the transformed mapping by $1*1$ convolutions. The outside solid lines  represent the identity mapping. The numbers in ResNet ``bottleneck" module and SEP-Net module represent the number of output channels  for illustration purpose.  }
 \label{fig:modules}
\end{figure*}

\begin{figure*}[]
  \centering
  \vspace{-0.1in}
   \includegraphics[scale=0.36]{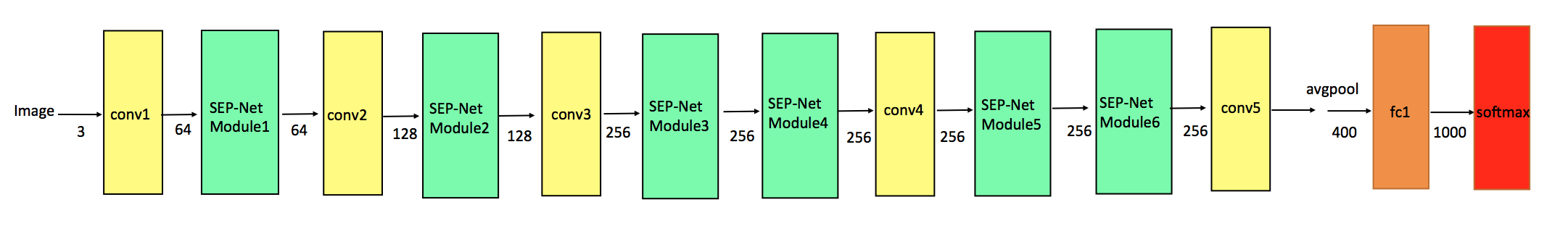}
  \caption{The architecture of our experimented SEP-Nets (the numbers illustrate the number of input/output  channels of each block for one SEP-Net with 1.7M parameters)}
  \label{fig:experimented-sep}
 \end{figure*}

\paragraph{SEP-Net Module and SEP-Net.} Built on the PRB, we design a new module for our SEP-Net, which is shown in  Figure \ref{fig:modules}(f). Our SEP-Net module consits of a dimension reduction layer ($1\times 1$ convolutions), 2 PRB blocks with different output channles, and a dimension recovery layer ($1\times 1$ convolutions). The last recovery layer enables us to add the skip connection as in ResNets, which is helpful for building up more layers. 
We also compare with other module designs of different networks in  Figure \ref{fig:modules}. In particular, comparing with the Fire module of the SqueezeNet, the SEP-Net module does not use the filter concatenation as introduced in Inception Net.  Comparing with the ResNet bottleneck module, we replace the $3\times 3$ convolutional layers with the PRB blocks. Finally, we plot the architecture of our experimented SEP-Nets in Figure \ref{fig:experimented-sep}. 
The detailed information of our experimented SEP-Nets (filter size, number of output channels, pad and stride of convolutional layers) is delayed to the supplement.  

\paragraph{Group-wise Convolution.} To further reduce the model size,  we adopt group convolution~\cite{Alexnet12} in our architecture. In particular, for each convolution (including both $1\times 1$ convolution and $3\times 3$ convolution), we split the input feature maps into $N$ groups and apply the corresponding convolution with a smaller number of channels to each group. The resulting feature maps in each group will be concatenated together.     
This simple design can effectively reduce the number of parameters by a factor of $N$. To see this, we can think about an example that maps an input feature map with $a$ channels to a feature map with $b$ channels. Using a single convolution $k\times k$, the size of the filter is $k*k*a*b$. If we use group convolution with $N$ groups, the size of each group filter is $k*k*\frac{a}{N}\times\frac{b}{N}$. As a result, the total size of all group filters is $k*k*a*b/N$.  In our implementation, we utilize group convolutions with a factor of 4 for all $1 \times 1$ and $3 \times 3$ kernels, thus reducing number of parameters by 4 times compared to that without using group convolutions. In Figure \ref{fig:group-convolution}(b), we show 4 group convolutions that are applied to all $1\times 1$ and $3 \times 3$ convolutions. It is notable that if we set the number of groups equal to the number of input channels, it degenerates to depth-wise convolutions as used in Google's MobileNets~\cite{howard2017mobilenets}. 


As a final note,  one could potentially train the pattern networks with binarized $k\times k$ filters from scratch, which might allow one to incorporate the domain knowledge into the design of binary patterns of $k\times k$ filters. Thus, it holds potential to benefit from  traditional used binary filters for designing deep but small neural networks. 

\section{Experiment}
\label{experimental-result}
In this section, we first present experimental results on CIFAR-10 \cite{krizhevsky2009learning} and ImageNet dataset \cite{Imagenet09} to justify that pattern binarization could reduce the effective number of parameters dramatically and fine-tuning other parameters of the binarized network  with fixed binarized pattern could achieve comparable performance to that of the original neural network models. Then, we show the designed \textbf{SEP-Net} structure could achieve better or comparable performance on ImageNet than using similar sized networks such as recently released Google MobileNets. We conduct all experiments using Caffe \cite{caffe14} open sourced library.
\subsection{CIFAR-10}
We first conduct experiments on the CIFAR-10 dataset \cite{krizhevsky2009learning}, which has 50,000 training images and 10,000 test images. Each image belongs to one of 10 classes and has RGB format with 32x32 size in the original data set.   
The data is preprocessed by Global Contrast Normalization (GCN) and ZCA whitening \cite{goodfellow2013maxout} and also padded by 4 pixels on each side of image. In the training phase, 32x32 crop is randomly sampled from the padded image while in the test phase we only test on the original image. We start the learning rate from 0.1 and divide by 10 at iteration 32k and 48k and the maximum number of iteration is 64K. The momentum is 0.9 and the weight decay is 0.0001. We train on one GPU using mini-batch SGD with a batch size 256. We report the test accuracy from the original paper and neural network models trained by us from scratch.  Note that the minor difference of performance between our results and that reported in the original paper might be due to data augmentation.  We apply pattern binarization   to several recent successful network structures: ResNet-20, ResNet-32, ResNet-44, ResNet-56 \cite{ResNet_cvpr16}. We first binarize 3x3 kernels in all convolutional layers of the obtained \textbf{Full} Model. For presentation purpose, we term the model after pattern binarization without fine-tuning as (\textbf{BiPattern} model). Then we do fine-tuning other parameters of \textbf{BiPattern} model fixing binarized pattern on the original Cifar10 data set to obtain \textbf{Refined} model. We report  test accuracy from the original paper denoted by \textbf{Ref}, and of the \textbf{Full} model, the \textbf{BiPattern} model and the \textbf{Refined} model in the Table \ref{cifar10-test-accuracy}. We can see from Table \ref{cifar10-test-accuracy}  that the test accuracy of \textbf{Refined} models are very close to that of \text{Full} models, 
which justify that 3x3 filters could be replaced with binary pattern extractions without sacrificing performance.  
\begin{table*}[]
\centering
\caption{Comparison on the-state-of-art models, \textbf{Full} model, \textbf{BiPattern} model, \textbf{Refined} model on \textbf{Cifar10} Dataset}
\label{cifar10-test-accuracy}
\begin{tabular}{l|l|llll}
\hline
Model    & Acc              &\textbf{Ref}             & \textbf{Full}   & \textbf{BiPattern}   & \textbf{Refined}                                           \\ \hline
ResNet-20 & \begin{tabular}[c]{@{}l@{}}Top-1\\ Top-5\end{tabular}  & \begin{tabular}[c]{@{}l@{}}0.9125\\       -\end{tabular} & \begin{tabular}[c]{@{}l@{}}0.9118\\ 0.9974\end{tabular} & \begin{tabular}[c]{@{}l@{}}0.1546\\ 0.5104\end{tabular} & \begin{tabular}[c]{@{}l@{}}0.8649\\ 0.9941\end{tabular} \\ \hline
ResNet-32 & \begin{tabular}[c]{@{}l@{}}Top-1\\ Top-5\end{tabular}  & \begin{tabular}[c]{@{}l@{}}0.9249\\      -\end{tabular}  & \begin{tabular}[c]{@{}l@{}}0.9276\\ 0.9972\end{tabular} & \begin{tabular}[c]{@{}l@{}}0.2634\\ 0.6932\end{tabular} & \begin{tabular}[c]{@{}l@{}}0.9021\\ 0.9962\end{tabular} \\ \hline
ResNet-44 & \begin{tabular}[c]{@{}l@{}}Top-1 \\ Top-5\end{tabular} & \begin{tabular}[c]{@{}l@{}}0.9283\\       -\end{tabular} & \begin{tabular}[c]{@{}l@{}}0.9283\\ 0.9982\end{tabular} & \begin{tabular}[c]{@{}l@{}}0.4825\\ 0.8765\end{tabular} & \begin{tabular}[c]{@{}l@{}}0.9145\\ 0.9965\end{tabular} \\ \hline
ResNet-56 & \begin{tabular}[c]{@{}l@{}}Top-1\\ Top-5\end{tabular}  & \begin{tabular}[c]{@{}l@{}}0.9303\\      -\end{tabular}  & \begin{tabular}[c]{@{}l@{}}0.9375\\ 0.9977\end{tabular} & \begin{tabular}[c]{@{}l@{}}0.5382\\ 0.9574\end{tabular} & \begin{tabular}[c]{@{}l@{}}0.9302\\ 0.9971\end{tabular} \\ \hline
\end{tabular}
\end{table*}
\\
\\
In Table \ref{cifar10-number-param} we compare the effective number of parameters between full ResNet-20, ResNet-32, ResNet-44, ResNet-56 models and their corresponding binarized pattern networks. The effective number of parameters of  the pattern network is referred to as the number of parameters that use floating point representations. In particular, we use one number to represent the binarized $3\times 3$ or $5\times5$ kernels by the scaling factor.  
From Table \ref{cifar10-number-param} we can see that the number of parameters  is reduced dramatically. For example the number of parameters in corresponding pattern network is reduced $86\%$ compared to original ResNet-56 model.

\begin{table}[]
\centering
\caption{Comparison of the number of parameters between \textbf{Full} networks and \textbf{Pattern} networks. We use one number to represent a binarized  $3\times3$ or $5\times5$ kernel.}
\label{cifar10-number-param}
\begin{tabular}{l|ll}
\hline
Model     & \textbf{Full Network} & \textbf{Pattern Network}  \\ \hline
ResNet-20 & 292K     & 55K             \\ \hline
ResNet-32 & 487K     & 78K             \\ \hline
ResNet-44 & 682K     &100K             \\ \hline
ResNet-56 & 876K     &123K             \\ \hline
\end{tabular}
\end{table}

\subsection{ImageNet}
\vspace{-.1in}
We also carry out experiments on the ImageNet 2012 classification data set \cite{Imagenet09}, which has 1.28 millions of training images and 50k validation images. Each image belongs to one of 1000 classes. We apply the same pattern binarization procedure used to ResNet on CIFAR-10 experiment 
to GoogLeNet \cite{googlenet15} and our customized Inception-net (denoted as C-InceptionNet) that removes all computationally expensive $5\times5$  convolutional kernels . For training GoogLeNet, we adopt the learning strategy from Caffe website and the learning rate follows a polynomial decay with the initial learning rate being $0.01$, momentum term $0.9$ and weight decay $0.0002$. We train GoogLeNet with a maximum number of iterations $600$K on one GPU and a batch size $128$. For training C-InceptionNet, we set initial learning rate as 0.1, divide learning rate 10 time after every 24 epochs, and train in total 90 epochs.  At the end, we compare test accuracy of the \textbf{Full} model, the \textbf{BiPattern} model, and the \textbf{Refined} model corresponding to GoogLeNet and C-InceptionNet. 

To illustrate the effect of binarizing different size filters, we report several results on GoogLeNet.  We first binarize $3\times 3$ filters fixing  $5 \times 5$ filters. Alternatively, we binarize $5 \times 5$ filters meanwhile fixing $3 \times 3$ filters. For comparison, we also conduct experiment by a) binarizing both $3 \times 3$ and $5 \times 5$ filters; b) binarizing only $1\times1 $ kernels. For C-InceptionNet, we only show the result of binarizing all $3\time 3$ filters. We present comparison results including test performance on \textbf{Refined} model using multicrop \cite{Alexnet12} in Table \ref{Imagenet-test accuracy}, from which we could see: i) for C-InceptionNet, the performance of the \textbf{Refined} model is competitive to the full model, just $0.8\%$ less than the full model on Top-1 accuracy; 
ii) for GoogLeNet, the performance of the \textbf{Refined} models of binarizing $k\times k$ filters ($k>1$) is significantly competitive to the full model; iii) binarizing $1\times1$ kernels suffers from more performance loss compared to binarizing $k\times k$ kernels ($k>1$), which justifies that our choice of binarizing $k \times k$ kernels ($k >1$) while retaining $1 \times 1$ kernels. All the performance numbers are based on a single center crop when performing testing  except for the last column,  which is included for future reference. 
\begin{table*}[]
\centering
\caption{Comparison on the-state-of-art models, \textbf{Full} model, \textbf{BiPattern} model, \textbf{Refined} model on \textbf{ImageNet} Dataset, including test accuracy from \textbf{Refined} model with multicrop. {\bf C-InceptionNet}: customized Inception-Net removing all $5\times 5$ convolutions.}
\label{Imagenet-test accuracy}
\begin{tabular}{l|l|lllll}
\hline
Model     & Acc         & \textbf{Ref} & \textbf{Full}  & \textbf{BiPattern}      & \textbf{Refined}  & \textbf{Multicrop}   \\ \hline
GoogLeNet & \begin{tabular}[c]{@{}l@{}}Top-1\\ Top-5\end{tabular} & \begin{tabular}[c]{@{}l@{}}-\\ 0.8993\end{tabular} & \begin{tabular}[c]{@{}l@{}}0.6865\\ 0.8891\end{tabular} & \begin{tabular}[c]{@{}l@{}}1x1 pattern:\\  0.0013\\ 0.0075\\ 
\hline{2-8}
3x3 pattern:\\ 0.3706\\ 0.6290\\ 

\hline
5x5 pattern:\\ 0.5141\\ 0.7619 \\ 
\hline
3x3 \& 5x5 pattern:\\ 0.1428\\ 0.31738\end{tabular} & \begin{tabular}[c]{@{}l@{}} \\0.6117 \\ 0.8395\\ \\0.6797\\ 0.8827\\ 
\\ 0.6917\\ 0.8904\\ \\ 0.6694\\ 0.8763\end{tabular} & \begin{tabular}[c]{@{}l@{}}\\0.636 \\0.856 \\ \\0.6893\\ 0.8898\\ \\ 0.6984\\ 0.8965\\ \\ 0.6812\\ 0.8844\end{tabular} \\ \hline

C-InceptionNet & \begin{tabular}[c]{@{}l@{}}Top-1\\ Top-5\end{tabular} &                                                    & \begin{tabular}[c]{@{}l@{}}0.648\\ 0.863\end{tabular} & \begin{tabular}[c]{@{}l@{}}0.0476\\ 0.1464\end{tabular}                                                                                                & \begin{tabular}[c]{@{}l@{}}0.6400\\ 0.8550\end{tabular}  & \begin{tabular}[c]{@{}l@{}}0.6521\\0.8626\end{tabular}                                                                                        \\ \hline
\end{tabular}
\end{table*}

In Table \ref{imagenet-num-parameter}, we compare the number of parameters of the full GoogLeNet and C-InceptionNet and their corresponding pattern networks. We can see that the pattern networks have dramatically less number of parameters that that of  the full models. For example through binarizing  $3\times3$ and $5\times5$ filters, we reduce the number of parameters by $44.6\%$ compared to that of the  full GoogLeNet.  

\begin{table*}[]
\centering
\label{imagenet-num-parameter}
\caption{Comparison of number of parameters between \textbf{Full} networks and \textbf{Pattern} networks. We use one number to represent a $3\times3$ or $5\times5$ kernel, resulting in parameter reduction.}
\begin{tabular}{l|llll}
\hline
Model                      & \textbf{Full Network}          & \textbf{Pattern Network} \\ \hline
\multirow{3}{*}{GoogLeNet} & \multirow{3}{*}{6.99M} & $3 \times 3$            & 4.43M                   \\ \cline{3-4} 
                           &                         & $5 \times 5$            & 6.43M                    \\ \cline{3-4} 
                           &                         & $3 \times 3$ \text{and} $5 \times 5$ &3.87M         \\ \hline
Customized-InceptionNet    &  5.10M           & \multicolumn{2}{r}{    2.43M}    \\ \hline
\end{tabular}
\end{table*}
\subsection{Comparison with The-State-Of-Art}
Finally, we compare the proposed \textbf{SEP-Net} with the SqueezeNet and the recently released MobileNet in terms of model size and classification accuracy. We present the performance of several variations of our SEP-Net architectures: SEP-Net-R, SEP-Net-B, SEP-Net-BQ. Here SEP-Net-R is our SEP-Net with raw filters. SEP-Net-B denotes SEP-Net with pattern binarization. SEP-Net-BQ further quantizes all other parameters to 8 bits.

As shown in Table \ref{tb:sepnet-mobilenet}, we  have trained two extremely small SEP-Nets suited for mobile/embedded devices. One model has 1.3M parameters, while the other has 1.7M parameters. The two SEP-Nets share the same neural network structure as Figure \ref{fig:experimented-sep}. The difference between the two SEP-Nets is: (1) in the SEP-Net with 1.7M parameters the last convolution layer uses a factor of 4 for group-wise convolution while the SEP-Net with 1.3M parameters uses a factor of 16 for the last layer group-wise convolution; (2) the output dimension of the last convolutional layers of the SEP-Net with 1.7M parameters  is 400 while that of the SEP-Net with 1.3M parameters is 512.  They produce $65.8\%$ and $66.7\%$ top 1 classification accuracy on the ImageNet dataset, respectively. Our small model outperforms MobileNet's equivalent with the same model size. Our larger model increases number of parameters by 0.3M, boosting the performance by $2\%$. 

For  easy comparison to other small CNNs, we also present the memory size of different models. Our \textbf{SEP-Net-B} reduces the model size to 4.2MB with slightly decreased accuracy of $63.7\%$ that equals to the performance of the MobileNet with 5.2MB. Our \textbf{SEP-Net-BQ} further reduces the storage or memory cost to $1.3$MB while maintaining roughly the same performance. It indicates that our extremely compact model also works with standard compression techniques.

\begin{table*}[]
\centering
\caption{Comparison between the designed \textbf{SEP-Net} and the MobileNet and the SqueezeNet in terms of model size and Top1 accuracy. {\bf SEP-Net-R}: SEP-Net with raw valued weights. {\bf SEP-Net-B}: SEP-Net with pattern binarization. {\bf SEP-Net-BQ}: SEP-Net with pattern binarization and other weights quantized using linear quantization with 8 bits. 
}
\label{tb:sepnet-mobilenet}
\begin{tabular}{l|ll|ll}
\hline
Model     & Parameter number       & Size (bytes)  & Top-1 Acc \\ \hline
MobileNet & \begin{tabular}[c]{@{}l@{}}1.3M\\ 2.6M\end{tabular} &  \begin{tabular}[c]{@{}l@{}}5.2MB\\ 10.4MB\end{tabular}   &  \begin{tabular}[c]{@{}l@{}}0.637\\ 0.684\end{tabular} \\ \hline
SEP-Net-R   & \begin{tabular}[c]{@{}l@{}}1.3M (small)\\ 1.7M (large)\end{tabular} &  \begin{tabular}[c]{@{}l@{}}5.2MB\\ 6.7MB\end{tabular}    &  \begin{tabular}[c]{@{}l@{}}  0.658\\ 0.667\end{tabular}\\ \hline \\ \hline

SqueezeNet~\cite{iandola2016squeezenet}  & 1.2M &  4.8MB    &  0.604\\
MobileNet~\cite{howard2017mobilenets} & 1.3M &  5.2MB   & 0.637\\
SEP-Net-R (Small)  & {\bf 1.3M} &  {\bf 5.2MB}    &  {\bf 0.658}\\
SEP-Net-B (Small)  & 1.1M &  4.2MB   & 0.637\\  
SEP-Net-BQ (Small)   & 1.1M &   1.3MB &  0.635  \\ \hline
\end{tabular}
\end{table*}

\section{Conclusion}
\label{conclusion}

In this paper, we view neural network operations as $1\times 1$ data transformation and $k\times k$ abstract pattern extraction. By converting $k\times k$ convolution kernels into binary patterns, we significantly reduced the model size as well as computational cost of modern neural network architectures such as InceptionNets and ResNets, without significantly sacrificing the network performances. Our binarization approach is extremely simple compared to previous literatures. We further proposed a small network architecture containing pattern residual blocks, which utilize binarized patterns to extract features and $1\times 1$ transformation to compute pattern residuals. 
The resulting concise neural network is small and effective compared to recent advances in compact neural network design. The effectiveness of our approach is demonstrated intensively on the CIFAR-10 dataset and the ImageNet dataset. We hope our investigation will inspire the community for advanced architecture design from a pattern point of view.   

{\small
\bibliographystyle{ieee}
\bibliography{ms}

\begin{thebibliography}{10}\itemsep=-1pt

\bibitem{cai2017deep}
Z.~Cai, X.~He, J.~Sun, and N.~Vasconcelos.
\newblock Deep learning with low precision by half-wave gaussian quantization.
\newblock {\em arXiv preprint arXiv:1702.00953}, 2017.

\bibitem{chen2015compressing}
W.~Chen, J.~Wilson, S.~Tyree, K.~Weinberger, and Y.~Chen.
\newblock Compressing neural networks with the hashing trick.
\newblock In {\em ICML}, 2015.

\bibitem{courbariaux2015binaryconnect}
M.~Courbariaux, Y.~Bengio, and J.-P. David.
\newblock Binaryconnect: Training deep neural networks with binary weights
  during propagations.
\newblock In {\em NIPS}, 2015.

\bibitem{courbariaux2016binarized}
M.~Courbariaux, I.~Hubara, D.~Soudry, R.~El-Yaniv, and Y.~Bengio.
\newblock Binarized neural networks: Training deep neural networks with weights
  and activations constrained to+ 1 or-1.
\newblock {\em arXiv preprint arXiv:1602.02830}, 2016.

\bibitem{Imagenet09}
J.~Deng, W.~Dong, R.~Socher, L.~Li, K.~Li, and L.~Fei-Fei.
\newblock Imagenet: a large-scale hierachical image database.
\newblock In {\em CVPR}, 2009.

\bibitem{gan2015devnet}
C.~Gan, N.~Wang, Y.~Yang, D.-Y. Yeung, and A.~G. Hauptmann.
\newblock Devnet: A deep event network for multimedia event detection and
  evidence recounting.
\newblock In {\em CVPR}, 2015.

\bibitem{fastrcnn15}
R.~Girshick.
\newblock Fast r-cnn.
\newblock In {\em ICCV}, 2015.

\bibitem{RCNN14}
R.~Girshick, J.~Donahue, T.~Darrell, and J.~Malik.
\newblock Rich feature hierarchies for accurate object detection and semantic
  segmentation.
\newblock In {\em CVPR}, 2014.

\bibitem{goodfellow2013maxout}
I.~J. Goodfellow, D.~Warde-Farley, M.~Mirza, A.~Courville, and Y.~Bengio.
\newblock Maxout networks.
\newblock {\em arXiv preprint arXiv:1302.4389}, 2013.

\bibitem{han2015deep}
S.~Han, H.~Mao, and W.~J. Dally.
\newblock Deep compression: Compressing deep neural networks with pruning,
  trained quantization and huffman coding.
\newblock {\em arXiv preprint arXiv:1510.00149}, 2015.

\bibitem{han2015adeep}
S.~Han, H.~Mao, and W.~J. Dally.
\newblock A deep neural network compression pipeline: Pruning, quantization,
  huffman encoding.
\newblock {\em arXiv preprint arXiv:1510.00149}, 10, 2015.

\bibitem{han2015learning}
S.~Han, J.~Pool, J.~Tran, and W.~Dally.
\newblock Learning both weights and connections for efficient neural network.
\newblock In {\em NIPS}, 2015.

\bibitem{he2017mask}
K.~He, G.~Gkioxari, P.~Doll{\'a}r, and R.~Girshick.
\newblock Mask r-cnn.
\newblock {\em arXiv preprint arXiv:1703.06870}, 2017.

\bibitem{ResNet_cvpr16}
K.~He, X.~Zhang, S.~Ren, and J.~Sun.
\newblock Deep residual learning for image recognition.
\newblock In {\em CVPR}, 2016.

\bibitem{howard2017mobilenets}
A.~G. Howard, M.~Zhu, B.~Chen, D.~Kalenichenko, W.~Wang, T.~Weyand,
  M.~Andreetto, and H.~Adam.
\newblock Mobilenets: Efficient convolutional neural networks for mobile vision
  applications.
\newblock {\em arXiv preprint arXiv:1704.04861}, 2017.

\bibitem{hubara2016binarized}
I.~Hubara, M.~Courbariaux, D.~Soudry, R.~El-Yaniv, and Y.~Bengio.
\newblock Binarized neural networks.
\newblock In {\em Advances in Neural Information Processing Systems}, pages
  4107--4115, 2016.

\bibitem{iandola2016squeezenet}
F.~N. Iandola, S.~Han, M.~W. Moskewicz, K.~Ashraf, W.~J. Dally, and K.~Keutzer.
\newblock Squeezenet: Alexnet-level accuracy with 50x fewer parameters and< 0.5
  mb model size.
\newblock {\em arXiv preprint arXiv:1602.07360}, 2016.

\bibitem{caffe14}
Y.~Jia, E.~Shelhamer, J.~Donahue, S.~Karayev, J.~Long, R.~Girshick,
  S.~Guadarrama, and T.~Darrell.
\newblock Caffe: Convolutional architecture for fast feature embedding.
\newblock {\em arXiv preprint arXiv:1408.5093}, 2014.

\bibitem{juefei2016local}
F.~Juefei-Xu, V.~N. Boddeti, and M.~Savvides.
\newblock Local binary convolutional neural networks.
\newblock {\em arXiv preprint arXiv:1608.06049}, 2016.

\bibitem{krizhevsky2009learning}
A.~Krizhevsky and G.~Hinton.
\newblock Learning multiple layers of features from tiny images.
\newblock 2009.

\bibitem{Alexnet12}
A.~Krizhevsky, I.~Sutskever, and G.~Hinton.
\newblock Imagenet classification with deep convolutional neural networks.
\newblock In {\em NIPS}, 2012.

\bibitem{li2016ternary}
F.~Li, B.~Zhang, and B.~Liu.
\newblock Ternary weight networks.
\newblock {\em arXiv preprint arXiv:1605.04711}, 2016.

\bibitem{lin2016overcoming}
D.~D. Lin and S.~S. Talathi.
\newblock Overcoming challenges in fixed point training of deep convolutional
  networks.
\newblock {\em arXiv preprint arXiv:1607.02241}, 2016.

\bibitem{lin2013network}
M.~Lin, Q.~Chen, and S.~Yan.
\newblock Network in network.
\newblock {\em arXiv preprint arXiv:1312.4400}, 2013.

\bibitem{lin2015neural}
Z.~Lin, M.~Courbariaux, R.~Memisevic, and Y.~Bengio.
\newblock Neural networks with few multiplications.
\newblock {\em arXiv preprint arXiv:1510.03009}, 2015.

\bibitem{long2015fully}
J.~Long, E.~Shelhamer, and T.~Darrell.
\newblock Fully convolutional networks for semantic segmentation.
\newblock In {\em CVPR}, 2015.

\bibitem{mellempudi2017ternary}
N.~Mellempudi, A.~Kundu, D.~Mudigere, D.~Das, B.~Kaul, and P.~Dubey.
\newblock Ternary neural networks with fine-grained quantization.
\newblock {\em arXiv preprint arXiv:1705.01462}, 2017.

\bibitem{ott2016recurrent}
J.~Ott, Z.~Lin, Y.~Zhang, S.-C. Liu, and Y.~Bengio.
\newblock Recurrent neural networks with limited numerical precision.
\newblock {\em arXiv preprint arXiv:1608.06902}, 2016.

\bibitem{rastegari2016xnor}
M.~Rastegari, V.~Ordonez, J.~Redmon, and A.~Farhadi.
\newblock Xnor-net: Imagenet classification using binary convolutional neural
  networks.
\newblock In {\em ECCV}. Springer, 2016.

\bibitem{ren2015faster}
S.~Ren, K.~He, R.~Girshick, and J.~Sun.
\newblock Faster r-cnn: Towards real-time object detection with region proposal
  networks.
\newblock In {\em NIPS}, 2015.

\bibitem{sermanet2013overfeat}
P.~Sermanet, D.~Eigen, X.~Zhang, M.~Mathieu, R.~Fergus, and Y.~LeCun.
\newblock Overfeat: Integrated recognition, localization and detection using
  convolutional networks.
\newblock {\em arXiv preprint arXiv:1312.6229}, 2013.

\bibitem{vggnet15}
K.~Simonyan and A.~Zisserman.
\newblock Very deep convolutional networks for large-scale image recognition.
\newblock In {\em ICLR}, 2015.

\bibitem{st2016fast}
P.-L. St-Charles, G.-A. Bilodeau, and R.~Bergevin.
\newblock Fast image gradients using binary feature convolutions.
\newblock In {\em CVPR}, 2016.

\bibitem{googlenet15}
C.~Szegedy, W.~Liu, Y.~Jia, P.~Sermanet, S.~Reed, D.~Anguelov, D.~Erhan,
  V.~Vanhoucke, and A.~Rabinovich.
\newblock Going deeper with convolutions.
\newblock In {\em CVPR}, 2015.

\bibitem{toshev2014deeppose}
A.~Toshev and C.~Szegedy.
\newblock Deeppose: Human pose estimation via deep neural networks.
\newblock In {\em CVPR}, 2014.

\bibitem{whitehill2006haar}
J.~Whitehill and C.~W. Omlin.
\newblock Haar features for facs au recognition.
\newblock In {\em Automatic Face and Gesture Recognition, 2006. FGR 2006. 7th
  International Conference on}, pages 5--pp. IEEE, 2006.

\bibitem{xu2016efficient}
Y.~Xu, H.~Yang, L.~Zhang, and T.~Yang.
\newblock Efficient non-oblivious randomized reduction for risk minimization
  with improved excess risk guarantee.
\newblock {\em arXiv preprint arXiv:1612.01663}, 2016.

\bibitem{xu2015discriminative}
Z.~Xu, Y.~Yang, and A.~G. Hauptmann.
\newblock A discriminative cnn video representation for event detection.
\newblock In {\em CVPR}, 2015.

\bibitem{yin2016training}
P.~Yin, S.~Zhang, J.~Xin, and Y.~Qi.
\newblock Training ternary neural networks with exact proximal operator.
\newblock {\em arXiv preprint arXiv:1612.06052}, 2016.

\bibitem{zha2015exploiting}
S.~Zha, F.~Luisier, W.~Andrews, N.~Srivastava, and R.~Salakhutdinov.
\newblock Exploiting image-trained cnn architectures for unconstrained video
  classification.
\newblock {\em arXiv preprint arXiv:1503.04144}, 2015.

\bibitem{zhou2016dorefa}
S.~Zhou, Y.~Wu, Z.~Ni, X.~Zhou, H.~Wen, and Y.~Zou.
\newblock Dorefa-net: Training low bitwidth convolutional neural networks with
  low bitwidth gradients.
\newblock {\em arXiv preprint arXiv:1606.06160}, 2016.

\end{thebibliography}
}

\section{Appendix}
We present the whole neural network structures for the designed \textbf{SEP-Net} with 1.7M parameters in Table \ref{label:sep-net-large-param} and \textbf{SEP-Net} with less parameters in Table \ref{label:sep-net-less-param}.

\begin{table*}[h]
\centering
\caption{The Neural Network Structure for the Designed \textbf{SEP-Net} with 1.7M parameters}
\label{label:sep-net-large-param}
\vspace{0.4cm}
\begin{tabular}{|l|l|l|l|l|}
\hline
Layer Type & \#Channel & Kernel size & Pad/Stride&\#Group  \\ \hline
conv1\_base           &64           &  $5\times5$           &1/2    & 1                       \\ \hline
st/sep-module1\_svd1\_base           &32           &  $1\times1$           &0/1 &1                           \\ \hline
st/sep-module1\_slice1\_1x1\_0\_base           &32           &  $1\times1$           &0/1&4                            \\ \hline
st/sep-module1\_slice1\_3x3\_0\_base           &32           &  $3\times3$           &1/1&4                            \\ \hline
st/sep-module1\_slice2\_1x1\_0\_base           &16           &  $1\times1$           &0/1&4                            \\ \hline
st/sep-module1\_slice2\_3x3\_0\_base           &16           &  $3\times3$           &1/1&4                            \\ \hline
st/sep-module1\_svd2\_base           &64           &  $1\times1$           &0/1          &1                  \\ \hline
conv2\_base           &128           &  $3\times3$           &1/2           & 1                \\ \hline
st/sep-module2\_svd1\_base           &64           &  $1\times1$           &0/1 &1                           \\ \hline
st/sep-module2\_slice1\_1x1\_0\_base           &64           &  $1\times1$           &0/1&4                            \\ \hline
st/sep-module2\_slice1\_3x3\_0\_base           &64           &  $3\times3$           &1/1&4                            \\ \hline
st/sep-module2\_slice2\_1x1\_0\_base           &32           &  $1\times1$           &0/1&4                            \\ \hline
st/sep-module2\_slice2\_3x3\_0\_base           &32           &  $3\times3$           &1/1&4                  \\ \hline
st/sep-module2\_svd2\_base           &128           &  $1\times1$           &0/1    &1                        \\ \hline
conv3\_base           &256           &  $3\times3$           &1/2                   &4         \\ \hline
st/sep-module3\_svd1\_base           &128           &  $1\times1$           &0/1    &1                        \\ \hline
st/sep-module3\_slice1\_1x1\_0\_base           &128           &  $1\times1$           &0/1&4                            \\ \hline
st/sep-module3\_slice1\_3x3\_0\_base           &128           &  $3\times3$           &1/1&4                            \\ \hline
st/sep-module3\_slice2\_1x1\_0\_base           &64           &  $1\times1$           &0/1 &4                           \\ \hline
st/sep-module3\_slice2\_3x3\_0\_base           &64           &  $3\times3$           &1/1 &4                 \\ \hline
st/sep-module3\_svd2\_base           &256           &  $1\times1$           &0/1          &1                  \\ \hline
st/sep-module4\_svd1\_base           &128           &  $1\times1$           &0/1          &1                  \\ \hline
st/sep-module4\_slice1\_1x1\_0\_base           &128           &  $1\times1$           &0/1&4                            \\ \hline
st/sep-module4\_slice1\_3x3\_0\_base           &128           &  $3\times3$           &1/1&4                            \\ \hline
st/sep-module4\_slice2\_1x1\_0\_base           &64           &  $1\times1$           &0/1 &4                           \\ \hline
st/sep-module4\_slice2\_3x3\_0\_base           &64           &  $3\times3$           &1/1 &4                 \\ \hline
st/sep-module4\_svd2\_base           &256           &  $1\times1$           &0/1          &1                  \\ \hline
conv4\_base           &256           &  $3\times3$           &1/2                         &1   \\ \hline
st/sep-module5\_svd1\_base           &128           &  $1\times1$           &0/1          &4                  \\ \hline
st/sep-module5\_slice1\_1x1\_0\_base           &128           &  $1\times1$           &0/1&4                            \\ \hline
st/sep-module5\_slice1\_3x3\_0\_base           &128           &  $3\times3$           &1/1&4                            \\ \hline
st/sep-module5\_slice2\_1x1\_0\_base           &64          &  $1\times1$           &0/1  &4                          \\ \hline
st/sep-module5\_slice2\_3x3\_0\_base           &64           &  $3\times3$           &1/1 &4                 \\ \hline
st/sep-module5\_svd2\_base           &256           &  $1\times1$           &0/1          &1                  \\ \hline
st/sep-module6\_svd1\_base           &128           &  $1\times1$           &0/1          &1                  \\ \hline
st/sep-module6\_slice1\_1x1\_0\_base           &128           &  $1\times1$           &0/1&4                            \\ \hline
st/sep-module6\_slice1\_3x3\_0\_base           &128           &  $3\times3$           &1/1&4                            \\ \hline
st/sep-module6\_slice2\_1x1\_0\_base           &64           &  $1\times1$           &0/1 &4                           \\ \hline
st/sep-module6\_slice2\_3x3\_0\_base           &64           &  $3\times3$           &1/1 &4                 \\ \hline
st/sep-module6\_svd2\_base           &256           &  $1\times1$           &0/1           &1                 \\ \hline
conv5\_base           &400           &  $3\times3$           &1/2                       &4     \\ \hline
\end{tabular}
\end{table*}
\begin{table*}[h]
\centering
\caption{The Neural Network Structure for the Designed \textbf{SEP-Net} with 1.3M parameters}
\label{label:sep-net-less-param}
\vspace{0.4cm}
\begin{tabular}{|l|l|l|l|l|}
\hline
Layer Type & \#Channel & Kernel size & Pad/Stride&\#Group  \\ \hline
conv1\_base           &64           &  $5\times5$           &1/2 &1                           \\ \hline
st/sep-module1\_svd1\_base           &32           &  $1\times1$           &0/1 &  1                         \\ \hline
st/sep-module1\_slice1\_1x1\_0\_base           &32           &  $1\times1$           &0/1  &4                          \\ \hline
st/sep-module1\_slice1\_3x3\_0\_base           &32           &  $3\times3$           &1/1  &4                          \\ \hline
st/sep-module1\_slice2\_1x1\_0\_base           &16           &  $1\times1$           &0/1  &4                          \\ \hline
st/sep-module1\_slice2\_3x3\_0\_base           &16           &  $3\times3$           &1/1  &4                          \\ \hline
st/sep-module1\_svd2\_base           &64           &  $1\times1$           &0/1         &1                   \\ \hline
conv2\_base           &128           &  $3\times3$           &1/2               & 1            \\ \hline
st/sep-module2\_svd1\_base           &64           &  $1\times1$           &0/1  & 1                         \\ \hline
st/sep-module2\_slice1\_1x1\_0\_base           &64           &  $1\times1$           &0/1 &4                           \\ \hline
st/sep-module2\_slice1\_3x3\_0\_base           &64           &  $3\times3$           &1/1 &4                          \\ \hline
st/sep-module2\_slice2\_1x1\_0\_base           &32           &  $1\times1$           &0/1 &4                           \\ \hline
st/sep-module2\_slice2\_3x3\_0\_base           &32           &  $3\times3$           &1/1 &4                 \\ \hline
st/sep-module2\_svd2\_base           &128           &  $1\times1$           &0/1    &1                        \\ \hline
conv3\_base           &256           &  $3\times3$           &1/2           &4                 \\ \hline
st/sep-module3\_svd1\_base           &128           &  $1\times1$           &0/1    &1                        \\ \hline
st/sep-module3\_slice1\_1x1\_0\_base           &128           &  $1\times1$           &0/1 &4                            \\ \hline
st/sep-module3\_slice1\_3x3\_0\_base           &128           &  $3\times3$           &1/1 &4                          \\ \hline
st/sep-module3\_slice2\_1x1\_0\_base           &64           &  $1\times1$           &0/1  &4                          \\ \hline
st/sep-module3\_slice2\_3x3\_0\_base           &64           &  $3\times3$           &1/1  &4                \\ \hline
st/sep-module3\_svd2\_base           &256           &  $1\times1$           &0/1        &1                    \\ \hline
st/sep-module4\_svd1\_base           &128           &  $1\times1$           &0/1         &1                   \\ \hline
st/sep-module4\_slice1\_1x1\_0\_base           &128           &  $1\times1$           &0/1&4                            \\ \hline
st/sep-module4\_slice1\_3x3\_0\_base           &128           &  $3\times3$           &1/1 &4                           \\ \hline
st/sep-module4\_slice2\_1x1\_0\_base           &64           &  $1\times1$           &0/1 & 4                          \\ \hline
st/sep-module4\_slice2\_3x3\_0\_base           &64           &  $3\times3$           &1/1 & 4                \\ \hline
st/sep-module4\_svd2\_base           &256           &  $1\times1$           &0/1        &  1                  \\ \hline
conv4\_base           &256           &  $3\times3$           &1/2                        &4    \\ \hline
st/sep-module5\_svd1\_base           &128           &  $1\times1$           &0/1           &1                  \\ \hline
st/sep-module5\_slice1\_1x1\_0\_base           &128           &  $1\times1$           &0/1 &4                           \\ \hline
st/sep-module5\_slice1\_3x3\_0\_base           &128           &  $3\times3$           &1/1 &4                          \\ \hline
st/sep-module5\_slice2\_1x1\_0\_base           &64          &  $1\times1$           &0/1   &4                       \\ \hline
st/sep-module5\_slice2\_3x3\_0\_base           &64           &  $3\times3$           &1/1  &4              \\ \hline
st/sep-module5\_svd2\_base           &256           &  $1\times1$           &0/1           &1                 \\ \hline
st/sep-module6\_svd1\_base           &128           &  $1\times1$           &0/1           &1                    \\ \hline
st/sep-module6\_slice1\_1x1\_0\_base           &128           &  $1\times1$           &0/1 &4                           \\ \hline
st/sep-module6\_slice1\_3x3\_0\_base           &128           &  $3\times3$           &1/1 &4                           \\ \hline
st/sep-module6\_slice2\_1x1\_0\_base           &64           &  $1\times1$           &0/1  &4                          \\ \hline
st/sep-module6\_slice2\_3x3\_0\_base           &64           &  $3\times3$           &1/1  &4                \\ \hline
st/sep-module6\_svd2\_base           &256           &  $1\times1$           &0/1           &1                    \\ \hline
conv5\_base           &512           &  $3\times3$           &1/2                   &16         \\ \hline
\end{tabular}
\end{table*}

\end{document}